\documentclass[runningheads]{llncs}
\usepackage{graphicx}
\usepackage{multirow}
\usepackage{todonotes}
\usepackage{xcolor}
\usepackage{adjustbox}
\usepackage{amsmath,amssymb}
\usepackage{amssymb}

\begin{document}
\title{Additive Angular Margin for Few Shot Learning to Classify Clinical Endoscopy Images}
\titlerunning{Additive Angular Margin for Few Shot Learning}
 \author{Sharib Ali\inst{1} \and
Binod Bhattarai\inst{2} \and
 Tae-Kyun Kim\inst{2} \and 
 Jens Rittscher\inst{1}}
 \authorrunning{S. Ali et al.}
 \institute{Institute of Biomedical Engineering, Big Data Institute, University of Oxford, UK\and Imperial College London, UK}
%
\maketitle              

\begin{abstract}

Endoscopy is a widely used imaging modality to diagnose and treat diseases in hollow organs as for example the gastrointestinal tract, the kidney and the liver. However, due to varied modalities and use of different imaging protocols at various clinical centers impose significant challenges when generalising deep learning models. Moreover, the assembly of large datasets from different clinical centers can introduce a huge label bias that renders any learnt model unusable. Also, when using new modality or presence of images with rare patterns, a bulk amount of similar image data and their corresponding labels are required for training these models. In this work, we propose to use a few-shot learning approach that requires less training data and can be used to predict label classes of test samples from an unseen dataset. We propose a novel additive angular margin metric in the framework of prototypical network in few-shot learning setting. We compare our approach to the several established methods on a large cohort of multi-center, multi-organ, and multi-modal endoscopy data. The proposed algorithm outperforms existing state-of-the-art methods.
\end{abstract}
%
\section{Introduction}

\begin{figure}[t!]
    \centering
    \includegraphics[trim=1cm 1cm 1cm 1cm,width=0.92\linewidth]{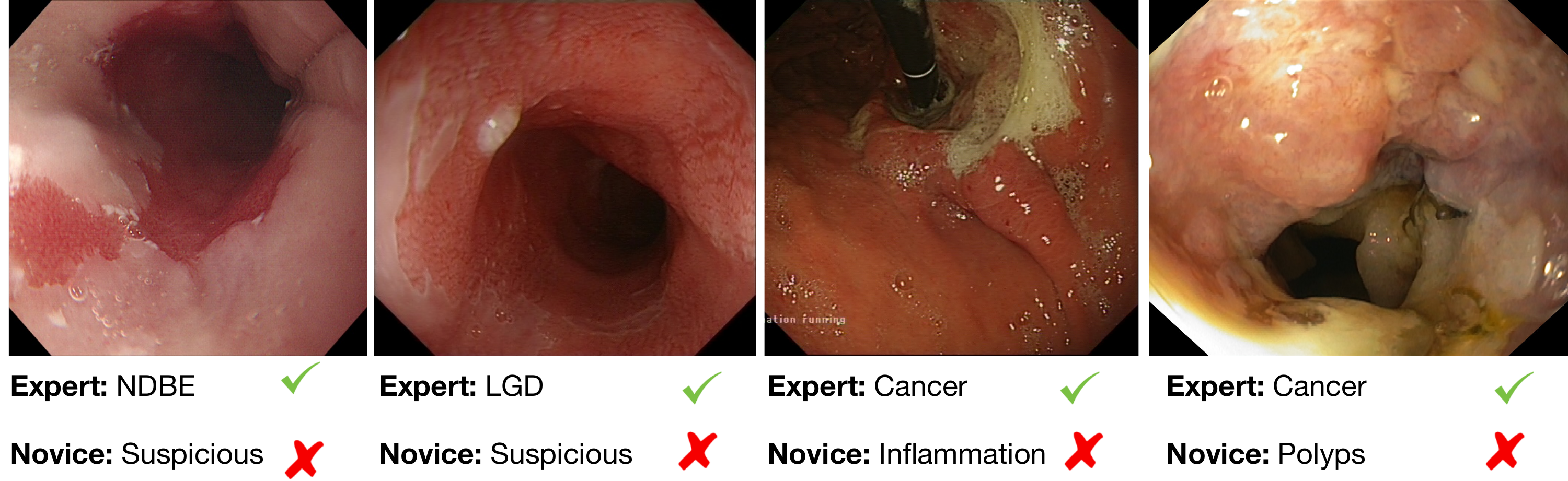}
    \caption{\textbf{Expert vs Novice.}  Examples of comparing expert ($>20$ years experience) labels with novice ($<3$ years experience) for some subtle cases in our dataset. Non-dysplastic (NDBE) and dysplastic (LGD) Barrett's are both graded as suspicious by the novice practitioner for two cases on the left. Similarly, some of the less subtle cases such as cancer are miss labelled as well by the novice annotator.}
    \label{fig:disease_motivation}
     \vspace{-0.25cm}
\end{figure}

The use of deep and non-linear models for medical applications is exponentially growing~\cite{biswas2019state}. The major bottleneck in training such data voracious models is the lack of abundant (expert) labeled data. Due to the requirement of large volume of labels, the possibility of surging incorrect labels is another problem. It is even more challenging on endoscopy data and in overall other similar medical domains due to: 1) difficulty to get domain experts to perform annotations, and 2) huge variability between expert and novice/trainee annotations (see Fig.~\ref{fig:disease_motivation}). The lack of publicly available datasets as well as their quality (\textit{e.g.}, missing and erroneous labels) pose additional challenges. Supporting new imaging protocols, such as new fluorescent imaging for endoscopy, is a case in which the entire training data would need to be collected, curated and annotated from scratch and in larger numbers. 

Given a training dataset with a large number of labeled samples we consider the case where one particular class with clinically relevant pattern is severely underrepresented. In this context, it is extremely challenging to train a model that will achieve an acceptable accuracy for such underrepresented class due to over-fitting. 
To overcome this problem, we propose to exploit the available annotated examples from related existing classes and use it to learn a metric parameter that can map test samples of an underrepresented and new class/classes without requiring large number of images and labels. 
Few-shot learning methods~\cite{vinyals2016matching,snell2017prototypical} which have been successfully used in the context of natural images, aim to
learn a model with a very small amount of labelled data. Some of the notable computer vision applications are in 
semantic segmentation~\cite{zhang2019canet}, object detection~\cite{wang2019few}, image segmentation~\cite{dong2018few}, and image-to-image translation~\cite{liu2019few}.

Here, we utilise this approach to overcome the problem of reliably detecting underrepresented classes. To date, few-shot learning has only been applied in a few selected medical image analysis applications. Mendela {\textit et. al}~\cite{medela2019few} utilised this technique to classify a set of 8 different histology patterns. However, their work lacks a systematic comparison of the state-of-the-art methods. Punch and colleagues~\cite{puch2019few} proposed to train Siamese network to minimise the triplet loss on Euclidean sub-space for brain imaging.
{}

\begin{figure}[t!]
    \centering
    \includegraphics[trim=1cm 1cm 1cm 1cm,width=0.92\linewidth]{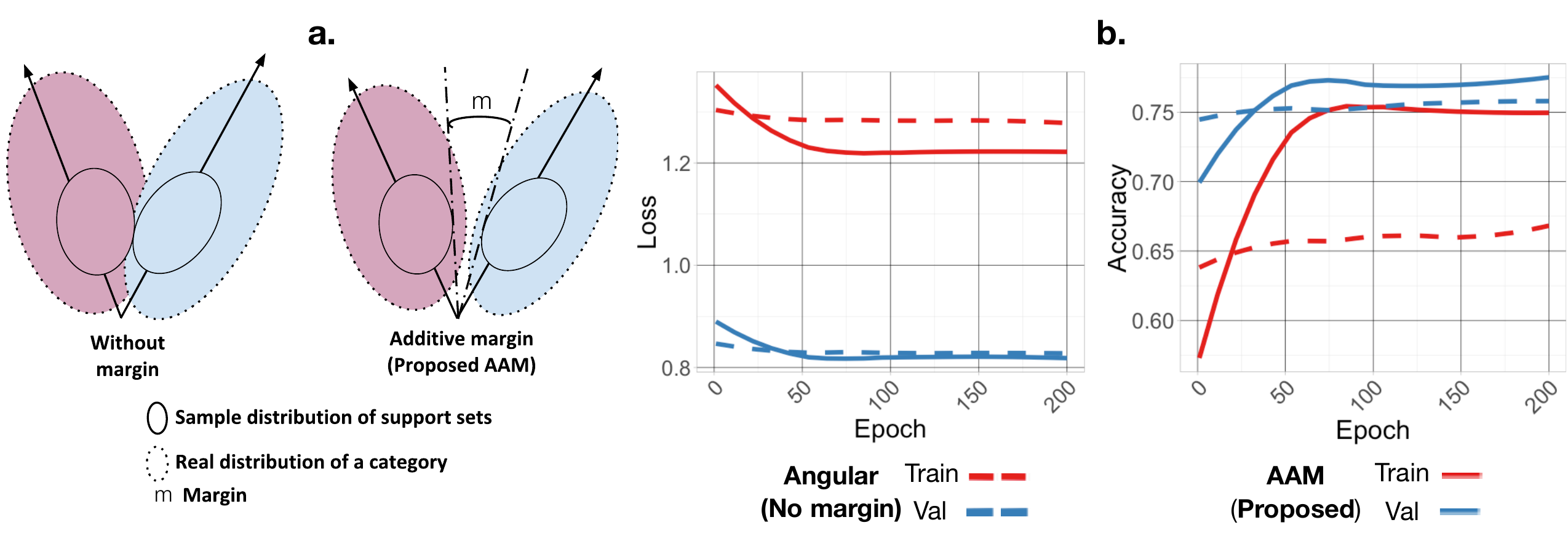}
    \caption{\textbf{Proposed Angular Additive Margin (AAM)}. a) Geometric interpretation of Soft max without margin (left) and with the margin (right). b) Loss and accuracy during training and validation with and without additive angular margin in (a).}
    \label{fig:additive_margin}
    \vspace{-0.25cm}
\end{figure}

Metric based few-shot learning algorithms have successfully been used in several tasks~\cite{vinyals2016matching,snell2017prototypical}. Studies have shown that the success of these approaches are due to the discriminative behaviour of the metric subspace~\cite{vinyals2016matching,snell2017prototypical,puch2019few}. Most existing methods rely on either minimising a triplet loss function~\cite{vinyals2016matching} or use of $\mathit{softmax}$ on Euclidean and subspace spanned by $\textit{cosine}$ function~\cite{vinyals2016matching,snell2017prototypical}. Qian {\it et al.}~\cite{qian2019softtriple} have recently demonstrated that the triplet loss is equivalent to using a $\mathit{softmax}$ loss. However, the $\mathit{softmax}$ loss is ineffective when number of classes becomes large and in presence of limited training examples. The drawback of these loss functions has been successfully addressed by incorporating an extra margin in metric subspace functions and was validated for face verification task~\cite{deng2019arcface,liu2017sphereface}. Adding margin in the metric-based loss function helps to generate more discriminative representations by widening the inter-class separation. Fig.~\ref{fig:additive_margin} shows the geometric interpretation of $\mathit{softmax}$ loss with and without the incorporated margin. Inspired by these ideas we introduce a margin on the objective function used in metric based few-shot learning~\cite{snell2017prototypical,vinyals2016matching} approach. To the best of our knowledge, this is the first work to introduce such an objective for few shot learning.
%
In this paper, we introduce a novel \textit{additive angular metric} (AAM) metric loss function in the prototypical network framework~\cite{snell2017prototypical} to perform few-shot learning based classification of diverse endoscopy samples that consists of multi-center, multi-modal (white light, narrow-band modality), underrepresented (e.g., low-grade dysplasia (LGD)), and hard samples (e.g., rare cancer samples, refer Fig.~\ref{fig:disease_motivation}). Fig.~\ref{fig:disease_motivation} illustrates cases which are differently labeled by an expert compared to a trainee: (Left) subtle images with NDBE and LGD, and (right) hard samples for cancer. This becomes more challenging when multi-center and multi-modality data are fused as in our case. We present extensive experiments for meta-training strategies and meta-testing performances on a diverse endoscopic data with {25 classes}. In summary, we make the following contributions:
\begin{itemize}
    \item Formulate a few-short learning approach to reliably classify multi-center, underrepresented, and hard samples in the context of endoscopy data
    \item Proposed a novel and generic angular distance metric for few-shot learning that improves the classification accuracy compared to the existing state-of-the-art methods
    \item An extensive evaluation of the proposed method and existing arts is done on the endoscopy benchmark
\end{itemize}{}
\section{Method}{\label{section:method}}

In this section, first we introduce few-shot learning and prototypical networks. Then, we introduce our proposed Additive Angular Margin
loss in the framework of such prototypical networks.\\
\noindent {\textbf{Few Shot Learning:}} Let us consider a scenario where there are two disjoint sets: 
$C_{train} \cap C_{test} = \emptyset$. There are a large number of training examples $(X_{train}^{N}, Y_{train}^{N})$ for the 
classes belonging to $C_{train}$ and only few training examples are available for the classes in $C_{test}$.
In medical domains, we can derive analogies  between these two sets with the well studied resource that have rich class categories compared to underrepresented classes that are clinically relevant patterns such as low-grade dysplasia in gastrointestinal endoscopy (see Fig.~\ref{fig:disease_motivation}). To enable model to generalise with such limited training examples, few shot learning algorithm can be used to train the model in an episodic paradigm.
In an episode trained in \textbf{$n$-shot and $k$-way manner},
$n$ training examples belonging to $k$ classes are used to learn the parameters of the model. 
Here, $n << N$ is a very small number of randomly sampled examples from the training set without replacement from a sub set of classes $n\times k << |C_{train}|$. These subset of annotated examples are 
called support set $\mathbb{S}_b$. At the same time, from the same sub-set of classes, we set aside few examples, not present in $\mathbb{S}_b$, as query set $\mathbb{Q}_b$. These are equivalent to the 
validation examples on supervised learning algorithms. However, the examples on query set $\mathbb{Q}_b$ changes in every episode, \textit{i.e.,} with the change in 
support sets $\mathbb{S}_b$. The model parameters $\phi$ are learnt $f_\phi:\mathrm{R}^D\rightarrow \mathrm{R}^d$ ($D >> d$) to make the correct predictions on the query set $\mathbb{Q}_b$. This procedure is repeated until the model selection criteria is met which is the accuracy on query set $\mathbb{Q}_b$ (for details please refer~\cite{wang2019few}). \\
\noindent{\textbf{{Prototypical} network~\cite{snell2017prototypical}:}} This few-shot learning model is simple yet obtains state-of-the-art performance
on several natural image benchmarks. The network computes a $d$-dimensional prototype representation $\phi \in \mathrm{R}^{d}$ for every class $k$ from 
the representations of support set $\mathbb{S}_b$ as shown in Eq.~(\ref{eqn:compute_prototype}). It then calculates distance $d(.)$ between the query example $X$ and the prototype representations $\phi$ of any class $k$ (see Eq.~\ref{eqn:dist_func})  and assign probability as shown in the Eq.~(\ref{eqn:soft_max_no_margin}). 
%
\begin{equation}
 L_k = \frac{1}{|S_b^{k}|}\sum_{(\textbf{x}_i, Y_i)\in S_b^{k}}f_\phi(\textbf{x}_i) 
 \label{eqn:compute_prototype}
 \vspace{-0.15cm}
\end{equation}{}
\begin{equation}
d(f_\phi(X), L_k) =  \arccos  \frac{f_\phi(X).L_k}{|f_\phi(X)|.|L_k|}
\label{eqn:dist_func}
\vspace{-0.15cm}
\end{equation}{}
\begin{equation}
p_\phi(y_k|X) = \frac{e^{\cos(d(f_\phi(X), L_k))}}{\sum_{k^{'}} e^{\cos(d(f_\phi(X), L_{k'})}}, \quad k'\in[1,..., k]
\label{eqn:soft_max_no_margin}
\vspace{-0.15cm}
\end{equation}{}
\begin{equation}
J_\phi = -log\,p_\phi(y_k|X)
\label{eqn:obj_function}
\vspace{-0.15cm}
\end{equation}{}

%
%

We have evaluated on both commonly used distance metrics: Euclidean ($l_2$) and $\textit{cosine}$~\cite{snell2017prototypical}. Given the subtle differences or the continuous nature of disease progression $\textit{cosine}$ metric is more effective to our purpose (see Sec.~\ref{sec:experiments}). The parameters are estimated via stochastic gradient descent (SGD) to minimise average negative log likelihood (see Eq.~(\ref{eqn:obj_function})) of target label $y_k$ for the query example ($X\in Q_b$).

\begin{figure}[t!]
    \centering
    \includegraphics[trim=1cm 1cm 1cm 1cm,width=0.92\textwidth]{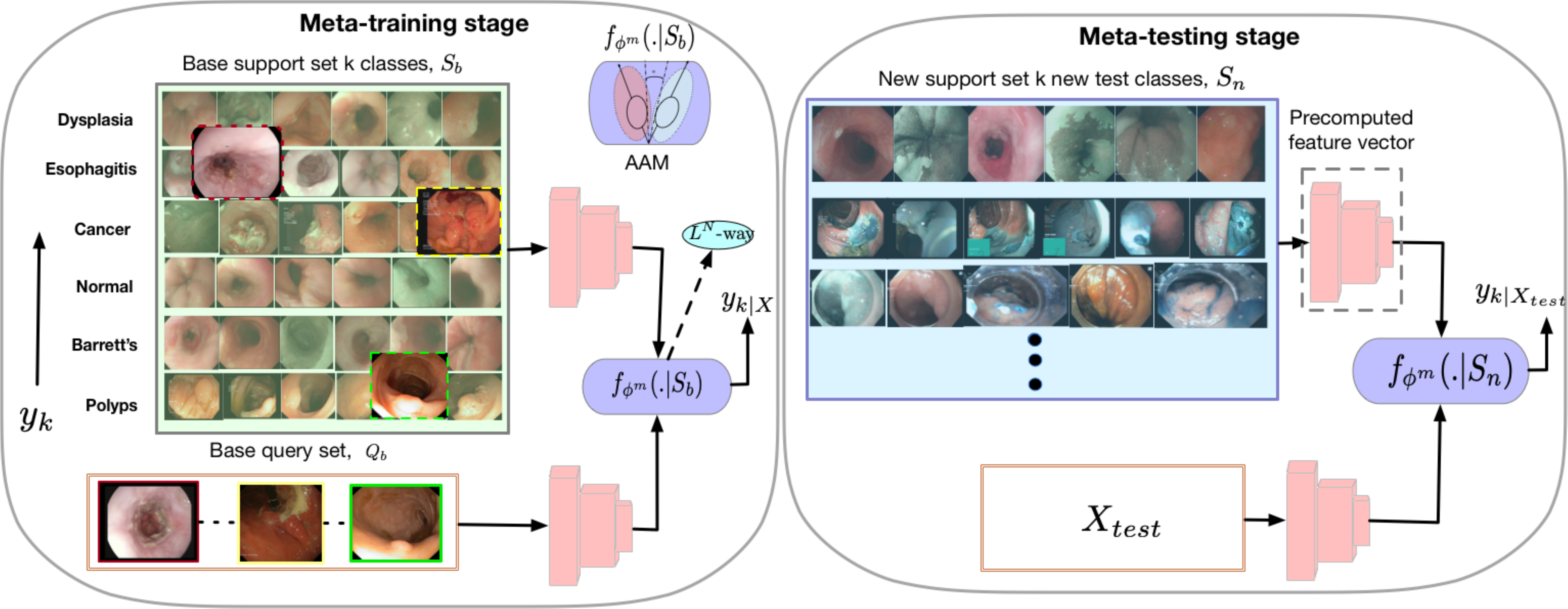}
    \caption{\textbf{Training and test strategies for few-shot classification for endoscopy.} (Left) Meta-training stage utilizes randomly generated sub-sets of base support and base query samples to learn the metric parameter $f(\phi)$ for which we propose an additive angular margin (AAM).}
    \label{fig:few-shot-paradigm}
     \vspace{-0.25cm}
\end{figure}{}
\noindent {\textbf{Proposed framework:}} Fig.~\ref{fig:few-shot-paradigm} shows schematic diagram of the proposed approach with two separate blocks outline the training and testing stages. Here, we utilise a prototypical network but introduce an additive angular margin on the $\mathbb{softmax}$ function. 

\textbf{Training stage:} There are two different sets of images: support set $\mathbb{S}_b$ and query set $\mathbb{Q}_b$ which we refer as base sets for our training stage. $\mathbb{S}_b$ consists of few randomly sampled images $n$ from a subset of $k$ classes. Embedding of these images computes their respective class prototypical representations $\phi$. These representations 
are compared with the query images $X^i\in Q_b$. Baseline prototypical model~\cite{snell2017prototypical} computes their probability $p_\phi$ of belonging to any class $k$ as in Eq.~(\ref{eqn:soft_max_no_margin}). As we argue that the limitations of such an approach can introduce restricted inter-class boundaries. To approximate the exact class distributions, we propose to incorporate an \textit{additive angular marginal} (AAM) with penalty margin $m$ as shown in Eq.~(\ref{eqn:soft_max_with_margin}). $m$ is an additive angular margin in radians and $\phi^m$ is the learnable parameter. This margin pushes samples apart which helps to approximate the real class distribution (see Fig.~\ref{fig:additive_margin} (a, right)). 
%
\begin{equation}
p_{\phi^m}(y_k|X) = \frac{e^{\cos(d(f_\phi(X), L_k)+m)}}{ e^{\cos(d(f_\phi(X), L_k)+m)} + \sum_{k', k' \ne k}  e^{\cos(d(f_\phi(X), L_{k'}))}}
\label{eqn:soft_max_with_margin}
\end{equation}{}
{
During training, only few $n$-train examples (e.g., $n=5$ in 5-shot) for $k$-classes (e.g., $k=5$ in 5-way) are used. Thus, requiring very small amount of data samples to learn the mapping function $\phi^m$. }
\textbf{Testing Stage:} Fig.~\ref{fig:few-shot-paradigm} (right) outlines the testing stage of the proposed approach. During inference time, we use the learnt $\phi^m$ to map both the support sets $S_n$ and query example $X_{test}$ to a embedding sub-space $f(\phi^m|S_n)$. It is to be noted that these class categories \textit{were not present in the training set}. From the embedding of the support set $S_n$ with $n$ examples for each class $k$, prototype representation are computed. Query examples $X_{test}^i$ are compared to the prototype representations of every classes $k^{test}$ which are later used to compute their distances. From these distances, first probability of target support class is computed and then the class with the highest probability $p_{{\phi}^m}$ is assigned to the target label.
%
%
\section{Experiments and results}
\label{sec:experiments}
\subsection{Dataset, training and evaluation metrics}
The dataset consists of \textbf{25 classes} with 60 images per class (1500 images) that has been collected from both in-house, international collaborators and online sources~\cite{KVASIR:2017,EDD2020} and consists of \textbf{multi-modal} (white light, narrow-band imaging and chromo-Endoscopy), \textbf{multi-center} (UK, France, Italy) and \textbf{multi-organ} (oesophagus, stomach, colon-rectum). We refer to this dataset as the \textit{miniEndoGI} classification dataset. The dataset consists of 8 classes from Kvasir~\cite{KVASIR:2017}, 5 classes from EDD2020~\cite{EDD2020}, and an additional 12 classes from our collected dataset from in-house and our collaborators (see \textbf{Supplementary materials Fig. 1}). Most of these classes are not publicly available and often found in scarce (e.g., gastric cancer or inflammation samples). 
The dataset was labelled independently by two expert endoscopists who have more than 20 years of experience. Since, the built dataset is from different sources, it captures a large variability in the clinical endoscopy imaging. Also, the use of the entire gastrointestinal tract (notably 3 organs: \textit{oesophagus, stomach, colon and rectum}) provides the evidence of robustness of our proposed approach. From this data cohort, we have used 15 classes for training, 5 for validation and 5 classes for testing. It is to be noted that \textit{no test class is present in the training set} and only sub-set of training data was used during meta-training stage (e.g., 1 sample for 1-shot and 5 samples for 5-shot).

The images were resized to $84 \times 84$  (similar to state-of-the-art methods that were applied on \textit{mini}ImageNet dataset). We have also used the four-block embedding architecture~\cite{vinyals2016matching,snell2017prototypical} for a fair comparison with other existing metric distances. Each baseline block consisted of 64-filter $3\times 3$ convolution, batch normalisation layer, a ReLU non-linearity and a $2\times 2$ max-pooling layer. The same encoding layer that produces $1024-$dimensional image features was used for both support and query samples (Fig.~\ref{fig:few-shot-paradigm}).
All models were trained for 200 epochs with 100 episodes per epoch and a learning rate (\textit{lr}) of $10^{-3}$ in SGD optimiser. We introduced a cutting rate of $1/3\times$ \textit{lr} for every 500 episodes. A stopping criteria was set if validation loss stops improving. For training, validation and test, same $n-$shot (samples) and $k-$way (classes) were chosen with set of 5 query samples for each experiment (see Sec.~\ref{section:method} for details).
%
%
\subsection{Quantitative evaluation}
%
\begin{table}[t!]
\centering
\begin{adjustbox}{width=1.0\textwidth}
\small{
\begin{tabular}{l|c|c|c|c|c|c|c}
\hline
\hline
\multirow{2}{*}{\textbf{Method}} & \multicolumn{1}{c|}{\multirow{2}{*}{\textbf{dist.}}} & \multicolumn{2}{c|}{\textbf{ 5-way }} & \multicolumn{2}{c|}{\textbf{ 3-way }} & \multicolumn{2}{c}{\textbf{2-way}} \\ \cline{3-8} 
 & \multicolumn{1}{c|}{} & \textbf{1-shot} & \textbf{5-shot} & \textbf{1-shot} & \textbf{5-shot} & \textbf{1-shot} & \textbf{5-shot} \\ \hline
\textbf{Baseline} & - & $34.40$ & $47.08$ & $47.53$ & $54.80$ & $52.59$   & $53.16$ \\ \hline
\textbf{Siamese} & Trip. loss~\cite{medela2019few} & $-$ & $-$ & $-$ & $-$ & $79.00\pm 4.24$   & $-$ \\ \hline
\multirow{3}{*}{\textbf{ProtoNet}} & Euclid.~\cite{snell2017prototypical} & $50.76\pm \textrm{\scriptsize{2.78}}$ & ${59.40}\pm \textrm{\scriptsize{1.55}}$ & $54.44\pm \textrm{\scriptsize{2.72}}$ & $73.80 \pm \textrm{\scriptsize{1.66}}$  & $68.80\pm \textrm{\scriptsize{1.51}}$ & $83.70\pm \textrm{\scriptsize{3.70}}$ \\ \cline{2-8} 
 & $\text{cosine}$~\cite{vinyals2016matching} & ${52.48}\pm \textrm{\scriptsize{1.95}} $ & $60.60\pm \textrm{\scriptsize{1.73}}$ &$73.67 \pm \textrm{\scriptsize{1.88}}$  &  $78.60\pm \textrm{\scriptsize{1.84}}$& $84.50\pm \textrm{\scriptsize{2.01}}$ & $88.80\pm \textrm{\scriptsize{1.77}}$ \\ \cline{2-8}    
 & AAM (our) & $\textbf{58.76}\pm \textrm{\scriptsize{1.64}} $ & $\textbf{66.72} \pm \textrm{\scriptsize{1.35}}$  &$\textbf{75.06}\pm \textrm{\scriptsize{1.87}}$  &  $\textbf{81.20}\pm \textrm{\scriptsize{1.72}}$& $\textbf{85.60}\pm \textrm{\scriptsize{2.21}}$ & $\textbf{90.60}\pm \textrm{\scriptsize{1.70}}$ \\ \hline
\end{tabular}
}
\end{adjustbox}
\caption{Few-shot classification using Conv-4 backbone on \textit{Endoscopy} dataset. Training was performed using $n$-shot, 5-query and $k$-way for every epoch and was tested on 5 new test class dataset, where $n=[1,5]$ and $k=[5, 3, 2]$. The entire training was done for 200 epochs with stopping criteria of minimum loss drop of $0.01$.
\label{tab:few-shot_baseline}
}
\vspace{-0.7cm}
\end{table}
Table~\ref{tab:few-shot_baseline} shows that the baseline CNN architecture poorly performs on small samples. However, when trained using prototypical network with metric learning approaches the classification performance improves significantly.
Notably, our proposed AAM surpasses all the state-of-the-art methods~\cite{snell2017prototypical}). We also have compared Siamese with triplet loss recently used technique in medical imaging~\cite{medela2019few}. The improvements over few-shot approaches are more than 6\% for 5-way and 1-shot, and 5-way and 5-shot. It is $>$2\% each for 3-way and $>$1\% for the 2-way classification cases for both 1-shot and 5-shot cases. Also, the standard deviation is the least for most cases which suggests that the proposed metric is able to well capture the inter-class variability present in the dataset. Nearly 80\%, and 90\% classification accuracies for 3-way, and 2-way settings shows a boost of more than 25\%, and 37\%, respectively, compared to the traditional deep-learning framework demonstrating the strength of our approach for classification of new classes and under representative samples on few samples (\textit{i.e.,} 1-shot referring to just 1 sample and 5-shot referring to just 5 samples). Additionally, the proposed AAM based prototypical network has an increased accuracy by nearly 7\% compared to the state-of-the-art Siamese network with triplet loss.

%
We also perform a comparative analysis on different network backbones shown in Table~\ref{tab:backbones}. It can be observed that our proposed method with embedded AAM gap on the $\textit{cosine}$ outperforms all the state-of-the-art metrics. There is a 2\% improvement for VGG-19, ResNet18, ResNet34 and DenseNet architectures compared to the baseline $\textit{cosine}$ metric. 
Fig.~\ref{fig:additive_margin}b demonstrates that the proposed AAM approach with additive margin minimises the categorical cross-entropy loss function better and have a large accuracy improvement over the classical $\textit{cosine}$ metric without the angular margin. As illustrated in the Fig.~\ref{fig:additive_margin}a, the learnt margin boosts the discriminating capability of the network.
%
\begin{table}[t!]
\centering
\begin{adjustbox}{width=0.9\textwidth}
\small{
\begin{tabular}{l|c|c|c|c|c|c}
\hline
\hline
\multicolumn{1}{c|}{\multirow{2}{*}{\textbf{Method}}} & \multicolumn{1}{c|}{\multirow{2}{*}{\textbf{dist.}}} & \multicolumn{5}{c}{\textbf{Network backbones}} \\ \cline{3-7} 
\multicolumn{1}{c|}{} & \multicolumn{1}{l|}{} & \textbf{VGG-19} &\textbf{ResNet18} & \textbf{ResNet34} & \textbf{ResNet50}  & \textbf{DenseNet} \\ \hline
\multirow{3}{*}{\textbf{ProtoNet}}
  & Euclid.~\cite{vinyals2016matching} & $20.00 \pm \textrm{\scriptsize{2.77}}$ &$64.40\pm \textrm{\scriptsize{1.35}}$ & $67.84\pm\textrm{\scriptsize{1.47}}$ & $62.60\pm \textrm{\scriptsize{1.21}}$  & $63.32\pm \textrm{\scriptsize{1.64}}$ \\ \cline{2-7} 
 & $\text{cosine}$~\cite{snell2017prototypical} & $22.04\pm \textrm{\scriptsize{1.70}}$ &$67.44\pm \textrm{\scriptsize{1.62}}$ & $67.44\pm \textrm{\scriptsize{1.86}}$ & $66.00\pm \textrm{\scriptsize{1.71}}$ & $ 69.00\pm \textrm{\scriptsize{1.44}}$ \\ \cline{2-7} 
 & AAM (ours) & $\textbf{24.16} \pm \textrm{\scriptsize{1.80}}$ &$\textbf{69.08}\pm \textrm{\scriptsize{2.18}}$ & $\textbf{70.00}\pm \textrm{\scriptsize{2.58}}$ & $\textbf{66.20} \pm \textrm{\scriptsize{2.91}}$ & $\textbf{71.00}\pm \textrm{\scriptsize{1.85}}$ \\ \hline
\end{tabular}
}
\end{adjustbox}
\caption{Few-shot classification on \textit{Endoscopyy} dataset for different backbone architectures. Networks were pretrained on \textit{ImageNet}. Test results are provided for 5-shot, 5-way, and 5-query.\label{tab:backbones}}
\vspace{-0.7cm}
\end{table}
\subsection{Qualitative evaluation}
\begin{figure}[t!]
    \centering
    \includegraphics[trim=1cm 1cm 1cm 1cm,width=0.95\textwidth]{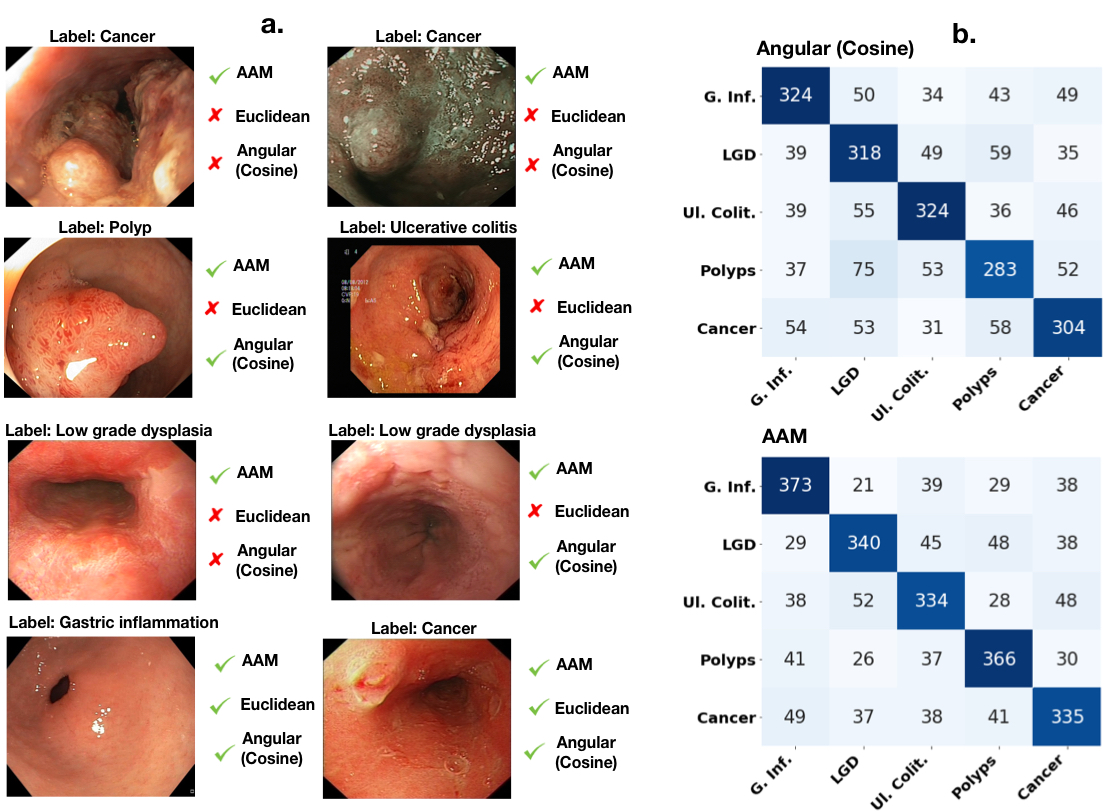}
    \caption{\textbf{Qualitative results on the test dataset for 5-shot, 5-way classification.} (a) Classification using different metric distances, and (b) confusion matrix showing outcome for each class for proposed AAM and angular ($\textit{cosine}$).}
    \label{fig:qualitative}
     \vspace{-0.5cm}
\end{figure}
From Fig.~\ref{fig:qualitative}a (1st row) it can be observed that `cancer' and `polyps' are confused by both Euclidean and angular metrics. This may be due to the protrusion seen in these images. While for the (2nd row), Euclidean metric miss classifies `cancer' and `ulcerative colitis' as `polyp'. Similarly, in Fig.~\ref{fig:qualitative}a (3rd row) low-grade dysplasia (LGD) in oesophagus was miss classified as cancer in both cases by Euclidean metric while $\textit{cosine}$ and the proposed method with AAM metric correctly identified them. For the  Fig.~\ref{fig:qualitative}a (4th row), due to the abundance of similar pyloric images for `gastric inflammation' (left) and a distinct protrusion with wound like structure (right), all methods correctly classified these labels. 
%
Fig.~\ref{fig:qualitative}b shows that the proposed AAM outperforms per class classification for all classes. It has nearly 20\% and 10\% improvement for `polyps' and `gastric inflammation' classes, respectively, compared with the $\textit{cosine}$ metric. Also, this confusion matrix suggests that most samples for `polyps' is confused with low-grade dysplasia (LGD) class and vice-versa for the $\textit{cosine}$ case which has been dealt widely by our novel AAM metric.  
The qualitative results in Fig.~\ref{fig:qualitative} shows that due to the added discriminative margin proposed in our AAM approach it can classify both hard and easy labels and do not confuse with more closer class types.
\section{Conclusion}
We have proposed a few-shot learning approach for classifying multi-modal, multi-center, and multi-organ gastrointestinal endoscopy data. In this context we proposed a novel additive angular metric in a prototypical network framework that surpasses other state-of-the-art methods. Our experiments on both traditionally used Conv-4 backbone and other widely used backbones showed an improved performance of our method in some cases by up to 7\% compared to recent works on few-shot learning approach. Finally, we have created a comprehensive endoscopy classification dataset that will be made publicly available to the community for further research in this area. In future, we plan to add more underrepresented classes in clinical endoscopy and explore other non-linear functions for improved inter-class separation.

\section*{Appendix}
\begin{figure}[b!]
    \centering
     \includegraphics[width=1\linewidth]{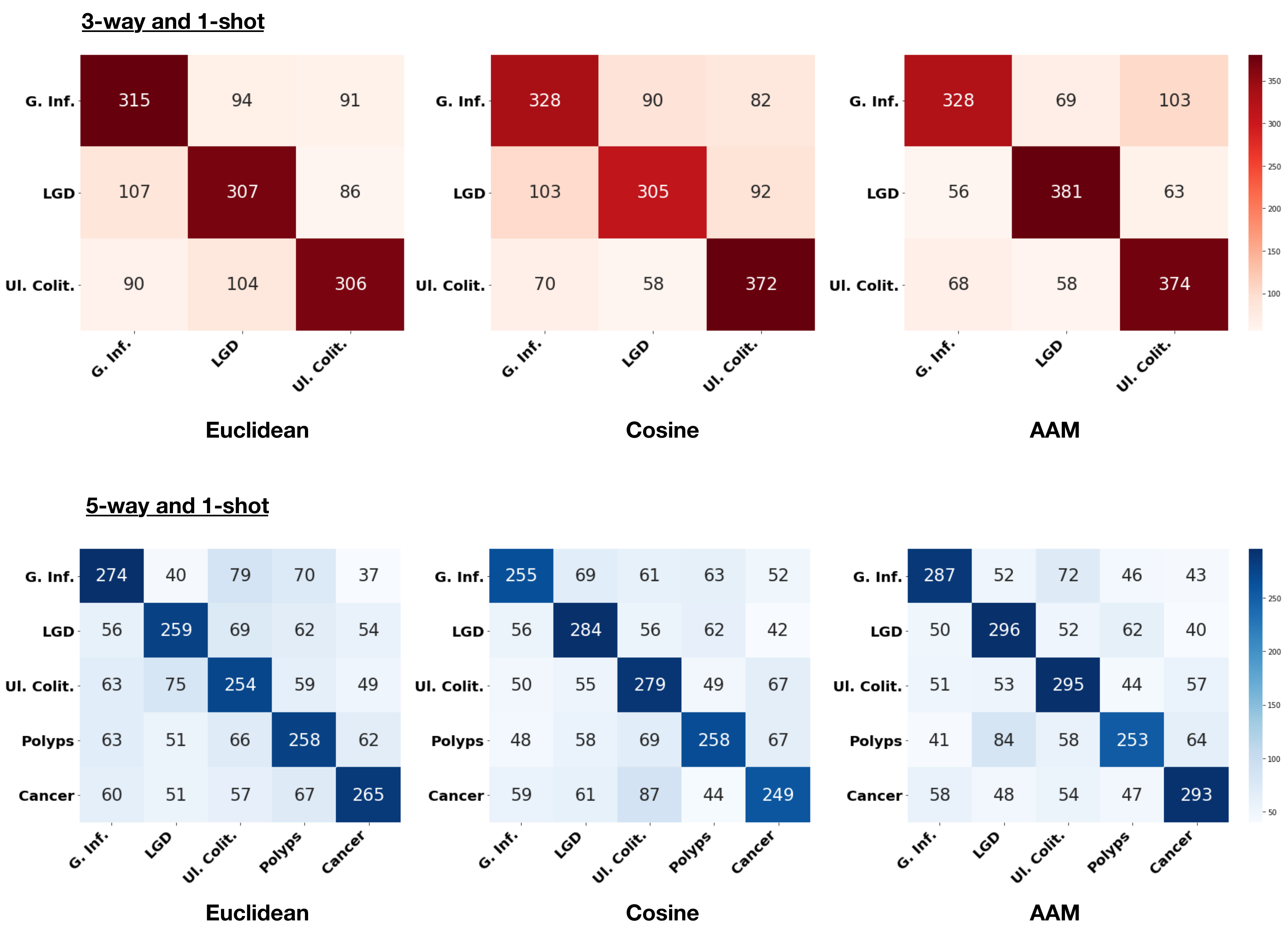}
      \caption{\textbf{Confusion matrix for 3-way and 5-way classification with a single sample (1-shot).} Top: 3-way and 1-shot, i.e., 1 sample was provided for classifying images from 3 classes ($k=3$). Bottom: 5-way and 1-shot, i.e., 1-sample was provided to predict between 5 classes ($k=5$). Random values $[1, 500]$ for $k$-test classes were provided for each case. Per class classification for each case is provided as the confusion matrix where the diagonal elements represents the correctly identified labels. Similar fashion was adapted for training on NVIDIA 2080Ti GPU. Please see Section 3.1 of the main manuscript regarding details on the training.}
    \label{fig:qualitative2}
\end{figure}

\paragraph{$k$-\textbf{way} \textbf{and} $1$-\textbf{shot}:}
We trained both 3-way and 5-way with 1-shot approach (\textit{i.e.}, only 1 sample was provided to train 3 class and 5 class classification) using state-of-the-art Euclidean and $\textit{cosine}$ distance based prototypical network using baseline Conv-4 backbone. The network was trained on NVIDIA 2080Ti which took nearly 3 hrs for training with an inference time of less than 0.01 ms. From Fig.~\ref{fig:qualitative2}, the proposed additive angular margin network provided significant increase in true positive for most cases compared to other prototypical networks.
\clearpage
\bibliographystyle{splncs04}
\bibliography{ref}
\end{document}